\documentclass[runningheads]{llncs}
\usepackage[T1]{fontenc}
\usepackage[nocompress]{cite}
\usepackage{amsmath,amssymb,amsfonts, dsfont}
\usepackage{graphicx}
\graphicspath{{Images/}{../figures/}}
\usepackage{float}
\usepackage{booktabs}
\usepackage{multirow}
\usepackage{textcomp}
\usepackage{xcolor}

\begin{document}
\title{GateSPINE: Gated Cross-View Fusion for Lumbar Spine MRI Report Generation
}
\titlerunning{Gated Cross-View Fusion for Lumbar Spine MRI Report Generation}
\author{Hoang Nguyen Van\inst{1} \and
Cuong Vuong Tuan\inst{1} \and
Trang Mai Xuan\inst{1}\thanks{Corresponding author.} \and
Bien Tran Van\inst{2,3} \and 
Nam Tran Van\inst{3} \and
Thien Van Luong\inst{4} 
}
\authorrunning{H. V. Nguyen et al.}
\institute{
Applied AI Lab, Phenikaa University, Hanoi, Vietnam\\
\and
Medical Imaging \& Radiological Technology Department, Faculty of Medical Technology, Phenikaa School of Medicine \& Pharmacy, Phenikaa University, Vietnam\\
\and
Radiology \& Functional Exploration Center, Phenikaa University Hospital, Vietnam
\\
\and
Business AI Lab, College of Technology, National Economics University, Vietnam\\
\email{23010101@st.phenikaa-uni.edu.vn, \{cuong.vuongtuan, trang.maixuan, bien.tranvan\}@phenikaa-uni.edu.vn, bv.namtv@phenikaamec.vn, thienlv@neu.edu.vn}
}
\maketitle 
\begin{abstract}
Automated report generation can ease the burden radiologists face when interpreting multi-sequence MRI studies. Unlike CT, MRI examinations comprise multiple sequences and imaging planes, each contributing complementary diagnostic information. Existing methods encode a study as a single volume and combine multiple acquisitions by fixed rules. Findings visible in only one plane are thus diluted and often missed, lowering recall on clinical efficacy metrics, where a missed abnormality is most costly. We propose GateSPINE, a vision-language framework that fuses sagittal T1 and T2 volumes with a training-free operator, encodes the fused sagittal and axial volumes with two parallel 3D encoders, and decodes their combined representation into a report. Its core mechanism is a gated cross-view fusion module that predicts, per feature channel and token, how much of each view to admit, so the more informative view dominates at each spatial location. We evaluate GateSPINE on three lumbar MRI datasets, comprising two public benchmarks and a private cohort collected from Phenikaa University Hospital, using both natural language generation (NLG) and clinical efficacy (CE) metrics. GateSPINE achieves the highest CE F1 through improved recall on all three datasets; on SPIDER, which lacks an axial sequence, this reflects the sagittal fusion component rather than the gated cross-view mechanism, which is validated on the two cohorts with both imaging planes. GateSPINE also remains competitive on standard NLG metrics.
\end{abstract}
\section{Introduction}
Magnetic resonance imaging (MRI) is widely used to evaluate degenerative lumbar spine diseases because it provides detailed visualization of soft tissues and neural structures~\cite{aaen2022nordsten,lafian2018olderadults}. In clinical practice, radiologists must examine multiple MRI sequences (e.g., T1W and T2W) acquired from different imaging planes (e.g., sagittal and axial), integrate the complementary findings across these images, and produce a comprehensive diagnostic report. This process is time-consuming because it requires careful interpretation of a large volume of imaging data, so automatic radiology report generation has attracted increasing attention~\cite{hamamci2024ct2rep, helmy2026spine,bai2024m3d}. 

Automatic report generation from CT typically couples a 3D vision encoder with a language decoder~\cite{hamamci2024ct2rep}. Most recently,
Reg2RG~\cite{chen2025reg2rg} grounds the report on anatomical regions and
combines region-level with global features, substantially improving diagnostic
accuracy. These methods assume that a study is a single volume, whereas an MRI
examination comprises several sequences and imaging planes, each showing some
findings more clearly than the others. 
For lumbar spine MRI specifically, SPINE~\cite{helmy2026spine}  stacks sagittal
T1- and T2-weighted images together with segmentation labels, showing that
integrating complementary contrasts improves report quality. Yet it models the
sagittal view alone. This omission is consequential because lateral recess
stenosis, facet hypertrophy, and the lateralization of disc herniation are assessed on axial images, findings that a sagittal-only model can therefore characterize only in part. CrossSpine~\cite{nguyen2026crossspine} attends across sequences to predict degeneration grades, but represents each by a single 2D slice. The two views are not interchangeable. Sagittal images trace continuity along the spinal axis and capture disc height, alignment, and marrow signal across all levels at once, whereas axial images expose a per-level geometry that sagittal slices cannot reconstruct. Which view carries the decisive evidence therefore changes from finding to finding, which a fixed combination rule cannot accommodate. To our knowledge, learning to combine sagittal and axial views for lumbar report generation remains unexplored.

To address this gap, we propose \textbf{GateSPINE}, a unified vision--language framework for lumbar MRI report generation. GateSPINE first
merges sagittal T1 and T2 with a training-free dynamic fusion operator, then encodes the fused sagittal and the axial volumes with two parallel 3D vision encoders. A lightweight gated cross-view fusion module learns, per feature channel, how much of each view to retain, and the resulting representation is decoded into a report by a vision--language model. In summary, our contributions are as follows:

\begin{itemize}
    \item We propose GateSPINE, a unified vision--language framework that integrates multi-sequence fusion, parallel 3D vision encoders, and adaptive cross-view fusion, improving both clinical-efficacy and language-generation performance.
  \item We design a lightweight gated cross-view fusion module that adaptively combines sagittal and axial representations. The module learns to emphasize the view that provides more informative evidence for each feature channel and spatial position, leading to more effective use of complementary information across MRI sequences and imaging planes.
    \item On the private PhenikaaMec cohort and two public datasets (SPIDER, Lumbar), GateSPINE improves clinical-efficacy micro-F1 over the strongest baseline by 3.6, 3.6, and 26.9 points, driven by recall gains of up to 38.0 points; on SPIDER, which lacks an axial sequence, this reflects the sagittal fusion component rather than the gated cross-view mechanism. GateSPINE also achieves the highest BERTScore on all three datasets (+0.3, +1.3, and +0.4 points).
\end{itemize}

\section{Related Work} \label{sect:related}
\subsubsection{CT report generation.}

Most 3D report generation systems extract volume features with a 3D vision encoder and pass them to a language decoder that writes the report. CT2Rep~\cite{hamamci2024ct2rep} is an early instance, though its decoder is trained from scratch, which limits fluency. Dia-LLaMA~\cite{chen2025diallama} replaces it with a pretrained LLM and prompts it with predicted disease labels, though it depends on an auxiliary classifier for those labels. Med3DVLM~\cite{xin2026med3dvlm} pursues a different axis, merging low- and high-level features to cut encoding cost, while generalist models treat report generation as one of many tasks~\cite{wu2025radfm,bai2024m3d}. Each of these systems therefore combines more than one feature stream. Yet all those streams are extracted from a single volume, so combining them never involves reconciling different views of the anatomy.

\subsubsection{MRI report generation.}
Report generation for MRI must contend with multiple sequences and imaging planes within one examination. SPINE~\cite{helmy2026spine} stacks sagittal T1 and T2 channel-wise with segmentation-derived channels, showing that combining contrasts improves report quality. Such stacking relies on both sequences being acquired in the same plane, so it does not extend to axial images, whose evidence the model never sees. Other systems take a different route, working from structured gradings~\cite{salem2025autospineai} or radiologist-authored findings~\cite{zanardo2026canai} rather than from the images themselves. How to combine sagittal and axial evidence therefore remains open. We address this by fusing sequences within a plane and gating across planes at the feature level.

\section{Methods}
\label{subsec:overview}

\begin{figure*}[t]
\centering
\includegraphics[width=\linewidth,trim={5 85 225 118},clip]{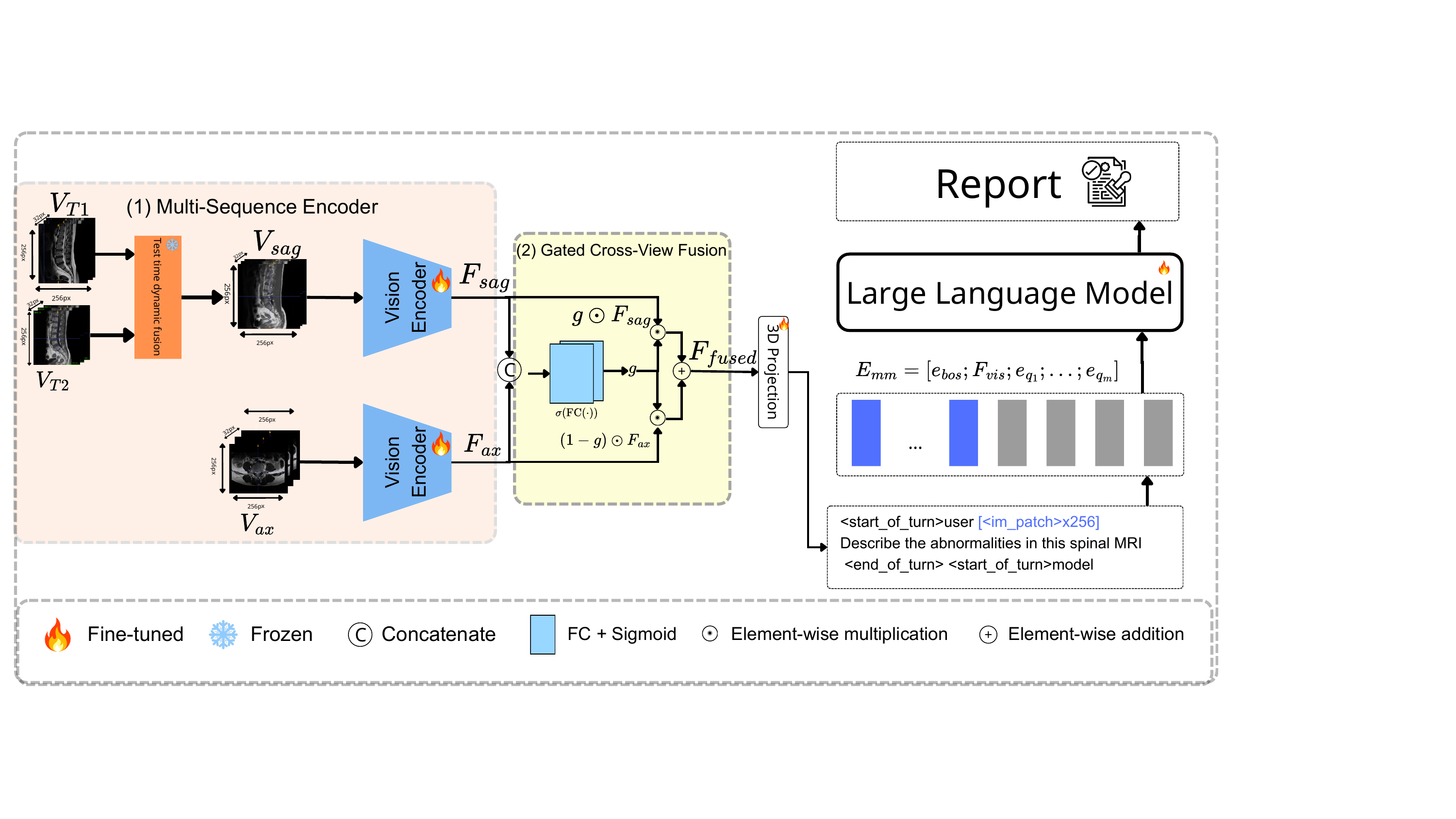}
\caption{Overview of the GateSPINE framework. Sagittal T1 and T2 volumes are fused by the TTD module, while the axial T2 volume is processed in parallel. Two ViT3D encoders extract features from the fused sagittal volume and the axial volume, and the Gated Cross-View Fusion layer combines them. The 3D Projector then compresses the fused features into visual tokens for the VLM decoder, which generates the final report.}
\label{fig:pipeline}
\end{figure*}

The overview of our method \textbf{GateSPINE} is shown in
Fig.~\ref{fig:pipeline}. Lumbar MRI report generation involves creating a report
$\mathcal{R}$ from a study of three volumes: sagittal T1-weighted
$\mathbf{V}^{\text{T1}}$, sagittal T2-weighted $\mathbf{V}^{\text{T2}}$, and
axial T2-weighted $\mathbf{V}^{\text{ax}}$, each in $\mathbb{R}^{D \times H
\times W}$. Unlike previous
works~\cite{hamamci2024ct2rep,wu2025radfm,xin2026med3dvlm} that encode a study as
a single volume, we preserve the two imaging planes as separate streams and let
the model decide, at each location, which to trust.

Since the sagittal sequences share a common geometry, we merge them into
$\tilde{\mathbf{V}}^{\text{sag}} = \varphi(\mathbf{V}^{\text{T1}},
\mathbf{V}^{\text{T2}})$ with test-time dynamic fusion (TTD)~\cite{cao2024test}, a training-free adaptive
fusion operator. Two parallel ViT3D encoders~\cite{dosovitskiy2021image}
$f_{\text{sag}}$ and $f_{\text{ax}}$, initialized from M3D~\cite{bai2024m3d},
then extract view-specific features. As the two planes may conflict locally, we
propose a gated cross-view fusion module $f_G$ that predicts, per channel and per
position, how much of each view to admit. A projector $f_P$ compresses the fused
features into visual tokens, which the MedGemma decoder~\cite{sellergren2025medgemma} consumes with a prompt $\mathbf{P}$ to
generate $\mathcal{R}$:
\begin{equation}
\mathcal{R} = \mathrm{LLM}\Big(f_P\big(f_G(\mathcal{F}_{\text{sag}},
\mathcal{F}_{\text{ax}})\big),\; \mathbf{P}\Big),
\label{eq:overview}
\end{equation}
where $\mathcal{F}_{\text{sag}} = f_{\text{sag}}(\tilde{\mathbf{V}}^{\text{sag}})$
and $\mathcal{F}_{\text{ax}} = f_{\text{ax}}(\mathbf{V}^{\text{ax}})$.
\subsection{Multi-Sequence Encoder}

\label{subsec:encoder}

This stage maps the three input volumes to two view-level representations $\mathcal{F}_{\text{sag}}$ and $\mathcal{F}_{\text{ax}}$: the two sagittal
sequences, which share an acquisition geometry, are merged before encoding, while the axial volume forms its own stream.

\noindent\textbf{Sagittal fusion.}
T1w and T2w are complementary, showing morphology and marrow signal versus fluid
and edema. Since a fixed rule cannot follow which contrast is locally stronger, we
merge them with TTD~\cite{cao2024test}, a training-free operator built on
CDDFuse~\cite{zhao2023cddfuse}. ``Test-time'' refers to the operator itself, which adapts its weights to each input without being trained on our data; we apply it identically to every study, as a fixed preprocessing step, both when training GateSPINE and at inference. Slice by slice, a shared pretrained autoencoder
reconstructs each source and measures its per-pixel squared error
$\ell^{s}(i,j)$, from which a \textit{Relative Dominability} weight follows:
\begin{align}
z^{s}(i,j) &= \exp\big(-\gamma\, \ell^{s}(i,j)\big), \\
w^{s}(i,j) &= \frac{\alpha \exp\big(z^{s}(i,j)\big)}
                   {\sum_{s' \in \mathcal{S}} \exp\big(z^{s'}(i,j)\big)},
\qquad s \in \mathcal{S},
\end{align}
where $\gamma > 0$ sets how sharply the weighting responds to error, and scaling
by $\alpha=|\mathcal{S}|$ makes the weights average one. A source reconstructing a location well
is thus emphasized while both are retained. The weighted features are decoded into
a fused slice, and stacking all slices gives $\tilde{\mathbf{V}}^{\text{sag}} =
\varphi(\mathbf{V}^{\text{T1}}, \mathbf{V}^{\text{T2}})$.

\noindent\textbf{View encoding.}
Both $\tilde{\mathbf{V}}^{\text{sag}}$ and $\mathbf{V}^{\text{ax}}$ are min--max
normalized and resampled to a common resolution $(D_t, H_t, W_t)$, then encoded
by two ViT3D encoders~\cite{dosovitskiy2021image} $f_{\text{sag}}$ and
$f_{\text{ax}}$ with separate weights, initialized from M3D~\cite{bai2024m3d}.
Each partitions its input into non-overlapping patches of size
$(p_D, p_H, p_W)$ and emits $N$ tokens of dimension $d$, so that
$\mathcal{F}_{\text{sag}}, \mathcal{F}_{\text{ax}} \in \mathbb{R}^{N \times d}$.
The axial branch requires no sequence fusion, as a single axial sequence is
acquired.

\subsection{Gated Cross-View Fusion}


\label{subsec:gcf}
This module takes the two view representations from Sec.~\ref{subsec:encoder} and
produces a single representation $\mathcal{F}_{\text{fuse}} \in
\mathbb{R}^{N \times d}$ in which each view contributes according to its content.

Concatenating the two token sets along the sequence would double their number to
$2N$ and raise the decoder cost accordingly, while averaging them applies the same
proportion at every channel and location. We instead combine them along the
\textit{feature} dimension with a learned gate, which leaves the token count
unchanged and lets the mixing proportion vary with content. Following the gated
fusion formulation of~\cite{arevalo2017gmu}, but predicting a gate per channel and
per token rather than one scalar per modality, we compute for each token
$n \in \{1, \dots, N\}$
\begin{align}
\mathbf{g}_n &= \sigma\!\big(\mathbf{W}_g\,
[\, \mathbf{f}^{\text{sag}}_n ;\; \mathbf{f}^{\text{ax}}_n \,]
+ \mathbf{b}_g\big) \;\in\; \mathbb{R}^{d}, \label{eq:gate} \\[2pt]
\mathbf{f}^{\text{fuse}}_n &= \mathbf{g}_n \odot \mathbf{f}^{\text{sag}}_n +
(\mathbf{1} - \mathbf{g}_n) \odot \mathbf{f}^{\text{ax}}_n, \label{eq:gatefuse}
\end{align}
where $\mathbf{f}^{\text{sag}}_n, \mathbf{f}^{\text{ax}}_n \in \mathbb{R}^{d}$ are
the $n$-th rows of $\mathcal{F}_{\text{sag}}$ and $\mathcal{F}_{\text{ax}}$,
$[\,\cdot\,;\,\cdot\,] \in \mathbb{R}^{2d}$ denotes vector concatenation,
$\mathbf{W}_g \in \mathbb{R}^{d \times 2d}$ and $\mathbf{b}_g \in \mathbb{R}^{d}$
are learnable, $\sigma(\cdot)$ is the sigmoid, and $\odot$ is element-wise
multiplication. Stacking over $n$ gives $\mathcal{F}_{\text{fuse}}$, at a cost of
$2d^2 + d$ parameters.

Because $\mathbf{g}_n$ is $d$-dimensional, different channels may be drawn from
different views within the same token, and the gate is conditioned on both views
jointly, so the proportion given to one depends on what the other contains. The
alternatives are recovered as special cases: a constant
$\mathbf{g}_n \equiv \tfrac{1}{2}$ gives averaging, and a gate held fixed across
tokens reduces to one scalar weight per view. 
\subsection{Vision--Language Projection and Decoding}
\label{subsec:proj}

This stage compresses $\mathcal{F}_{\text{fuse}}$ into visual tokens the language
model can consume. We reshape it into its 3D patch grid and apply non-overlapping
average pooling with factor $\rho$ per axis, giving $N_v = N/\rho^{3}$ tokens
$\mathcal{F}_{\text{seq}} \in \mathbb{R}^{N_v \times d}$ while preserving spatial
structure. Following the projection design of the pretrained
VLM~\cite{sellergren2025medgemma}, a trainable adapter maps these into its
$d_m$-dimensional space, after which its own frozen RMSNorm and projection
$\mathbf{W}_p$ are applied:
\begin{equation}
\mathcal{F}_{\text{vis}} = \mathbf{W}_p\, \mathrm{RMSNorm}\!\big(
\mathrm{GELU}(\mathbf{W}_a \mathcal{F}_{\text{seq}} + \mathbf{b}_a)\big)
\;\in\; \mathbb{R}^{N_v \times d_{\text{LLM}}}.
\label{eq:projection}
\end{equation}
Training only the adapter $(\mathbf{W}_a, \mathbf{b}_a)$ reuses the pretrained
vision--language alignment. The visual tokens then replace the placeholder
embeddings in $\mathbf{P}$, forming a multimodal sequence of length
$M = 1 + N_v + m$ from a BOS token, $N_v$ visual tokens, and $m$ prompt tokens.
We adapt the decoder with LoRA~\cite{hu2022lora} on all linear layers, while the
view encoders, the gate, and the adapter are trained with full gradients.
\subsection{Training and Inference}
\label{subsec:train}

The model is trained with an autoregressive cross-entropy loss over
report tokens $\mathcal{R} = \{r_1, \ldots, r_T\}$, conditioned on the multimodal
input:
\begin{equation}
\mathcal{L} = -\sum_{t=1}^{T} \mu_t \,
\log p_\theta\!\big(r_t \mid r_{<t}, \mathcal{F}_{\text{vis}}, \mathbf{P}\big),
\qquad \mu_t = \mathds{1}[\,r_t \text{ is an answer token}\,],
\label{eq:loss}
\end{equation}
where $\mu_t$ restricts the gradient to answer tokens, so the model is not
trained to reproduce the prompt. During training the prompt is sampled at random
from a set of clinical query templates, improving robustness to prompt wording.
At inference, reports are decoded greedily, stopping at the end-of-sequence token
or after $T_{\max}$ tokens.
\section{Experiments and Results} \label{sect:experiment}

\subsection{Datasets and Evaluation Metrics}

\textbf{Datasets.} PhenikaaMec~\cite{vu2026phenspines} is an internal cohort of 250 patients (112/138
M/F, aged $48.1\pm15.2$ years) with sagittal T1, sagittal T2, and axial T2 at $0.3\times0.3\times4.4$~mm, extracted from hospital PACS and anonymized under appropriate ethical clearance; 236 remain after discarding cases lacking a report or axial sequence. The reports are translated from Vietnamese with LLaMA-3.3-70B, and we split by acquisition month into 118/58/60 studies training on earlier months and testing on later ones to mirror deployment. Lumbar~\cite{alkafri2019boundary} contains 575 Mendeley cases (one per patient) with both planes; 515 remain after discarding empty reports, split 70:15:15. SPIDER~\cite{vandergraaf2024lumbar} provides sagittal T1, T2, and segmentation masks but no axial sequence and no free-text reports, split 60:20:20 by patient. On SPIDER, all methods use the reference reports released by SPINE~\cite{helmy2026spine}, generated from grading labels with GPT-4. Without an axial sequence the cross-view module is inactive, and, following SPINE, the mask is combined with T1 and T2 per slice in ratio $2\!:\!2\!:\!3$ in place of TTD;  SPIDER thus tests the sagittal branch, while the gated cross-view module is evaluated on the two cohorts with both planes.

\noindent\textbf{Evaluation Metrics.} Following recent report generation work~\cite{helmy2026spine, chen2025reg2rg}, we report language quality (NLP) and clinical efficacy (CE) metrics. The NLP metrics are BLEU-4~\cite{papineni2002bleu}, ROUGE-1, ROUGE-L~\cite{lin2004rouge}, METEOR~\cite{banerjee2005meteor}, and BERTScore~\cite{zhang2020bertscore}. These cannot capture diagnostic correctness, since a negation flips meaning while leaving surface similarity intact. CE metrics therefore compare clinical entities: an LLM judge (LLaMA-3.3-70B) labels nine lumbar findings---disc herniation, bulge, degeneration, spinal stenosis, osteophytes, nerve compression, curvature abnormality, annular tear, and disc height loss---as present or absent in both generated and reference reports. We report Micro-F1 pooled over findings and Macro-F1 averaged over them. The same judge scores every method, so the comparison is consistent across models; validating the judge and the report translation against radiologist annotation is left to future work.

\subsection{Implementation Details and Baselines}
\textbf{Implementation.} GateSPINE is implemented in PyTorch and trained on one NVIDIA Quadro RTX 8000, with MedGemma 4B-IT as decoder and both ViT3D encoders initialized from M3D~\cite{bai2024m3d}. Volumes are resampled to $D_t \times H_t \times W_t = 32 \times 256 \times 256$ and split into $(4,16,16)$ patches ($N=2048$, $d=768$); pooling with $\rho=2$ gives $N_v=256$ visual tokens, projected via $d_m=1152$ to $d_{\text{LLM}}=2560$, and TTD uses $\gamma=50$. Stage~1 trains the encoders, gate, and adapter with the LLM frozen (learning rate $10^{-4}$); stage~2 adds LoRA~\cite{hu2022lora} (rank 8, scaling 32, dropout 0.05) and fine-tunes jointly with a per-dataset learning rate. Both stages use AdamW with cosine decay, mixed precision, an effective batch size of 8 via gradient accumulation, and 10--15 epochs with early stopping.

\noindent\textbf{Baselines.} We compare against CT2Rep~\cite{hamamci2024ct2rep}, Med3DVLM~\cite{xin2026med3dvlm}, Dia-LLaMA~\cite{chen2025diallama}, and SPINE~\cite{helmy2026spine}, retrained on each dataset with the same preprocessed inputs and greedy decoding. Being single-volume methods, they receive the fused sagittal volume without the axial branch, so this comparison reflects the framework as a whole; the ablation in Sec.~\ref{sect:ablation}, where every variant receives both planes, isolates the fusion mechanism.

\subsection{Evaluation Results}
\begin{figure}[!t]
\centering
\includegraphics[width=0.45\columnwidth]{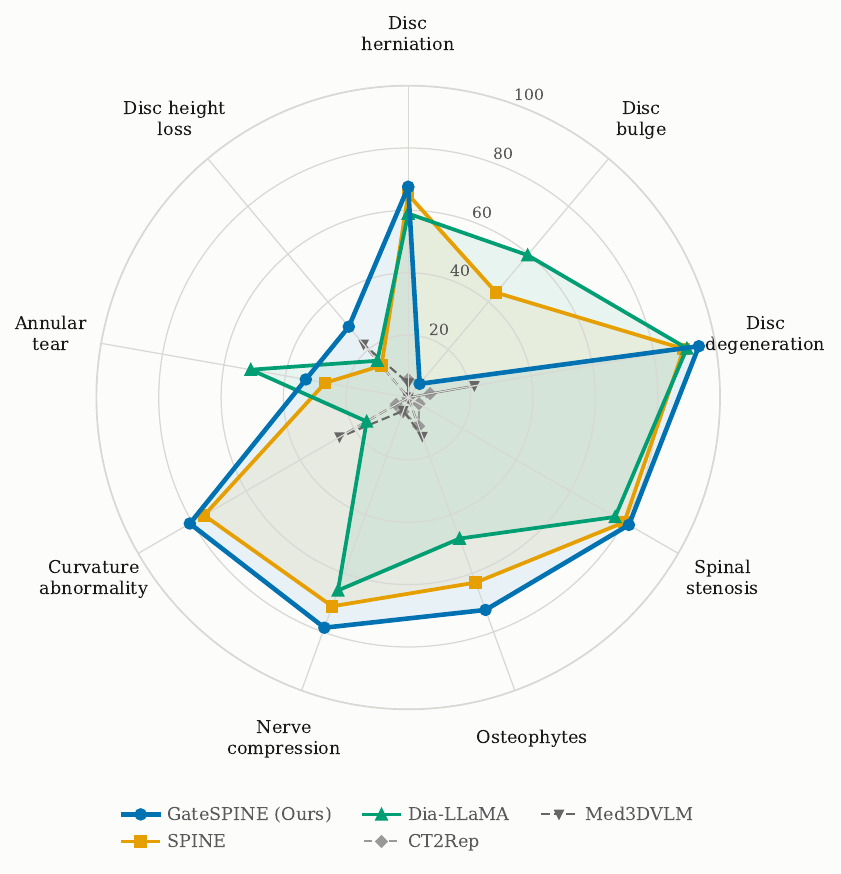}
\caption{F1 score for each clinical finding on PhenikaaMec.}
\label{fig:pathology_radar}
\end{figure}

\begin{table}[!t]
\centering
\caption{Comparison with state-of-the-art methods on language-quality (NLP) and clinical-efficacy (CE) metrics across three datasets. Best and second-best results in \textbf{bold} and \underline{underlined}.}
\label{tab:result_main}
\scriptsize
\setlength{\tabcolsep}{1.9pt}
\begin{tabular}{@{}c l | c c c c c | c c c | c c c@{}}
\hline\hline
& \multirow{2}{*}{\textbf{Method}} & \multicolumn{5}{c|}{\textbf{NLP}} & \multicolumn{3}{c|}{\textbf{CE Micro}} & \multicolumn{3}{c}{\textbf{CE Macro}} \\
\cmidrule(lr){3-7} \cmidrule(lr){8-10} \cmidrule(lr){11-13}
& & \textbf{BL-4} & \textbf{RG-1} & \textbf{RG-L} & \textbf{MTR} & \textbf{BERT} & \textbf{Pre.} & \textbf{Rec.} & \textbf{F1} & \textbf{Pre.} & \textbf{Rec.} & \textbf{F1} \\
\midrule
\multirow{5}{*}{\rotatebox{90}{\scalebox{0.8}{PhenikaaMec}}}
& CT2Rep~\cite{hamamci2024ct2rep}   & 25.91 & 57.97 & 44.21 & 45.58 & 89.78 & 66.67 & 2.37 & 4.58 & 61.11 & 2.08 & 3.99 \\
& Dia-LLaMA~\cite{chen2025diallama} & \textbf{28.16} & \underline{58.99} & \underline{45.05} & \textbf{50.57} & \underline{90.88} & \underline{74.79} & 52.05 & 61.38 & \textbf{71.79} & 47.75 & 53.55 \\
& Med3DVLM~\cite{xin2026med3dvlm}   & 4.35 & 36.99 & 20.27 & 22.78 & 87.03 & 36.51 & 6.69 & 11.30 & 28.86 & 6.68 & 10.19 \\
& SPINE~\cite{helmy2026spine}       & 25.04 & 56.84 & 42.71 & 45.99 & 90.43 & 71.24 & \underline{64.02} & \underline{67.49} & 65.23 & \underline{57.17} & \underline{58.78} \\
\cmidrule(lr){2-13}
& \textbf{GateSPINE}                & \underline{26.46} & \textbf{60.24} & \textbf{46.15} & \underline{48.63} & \textbf{91.16} & \textbf{75.33} & \textbf{67.35} & \textbf{71.12} & \underline{65.30} & \textbf{59.69} & \textbf{60.50} \\
\midrule
\multirow{5}{*}{\rotatebox{90}{SPIDER}}
& CT2Rep~\cite{hamamci2024ct2rep}   & 62.76 & \underline{77.13} & \underline{77.37} & \underline{87.45} & \underline{94.83} & 87.50 & 50.60 & 64.12 & 59.17 & 40.46 & 47.24 \\
& Dia-LLaMA~\cite{chen2025diallama} & 53.79 & 64.70 & 63.59 & 64.35 & 93.65 & \underline{89.47} & \underline{76.12} & \underline{82.26} & \underline{70.99} & \underline{61.67} & \underline{64.98} \\
& Med3DVLM~\cite{xin2026med3dvlm}   & 45.87 & 59.81 & 58.69 & 57.79 & 93.06 & \textbf{93.55} & 68.24 & 78.91 & \textbf{84.96} & 56.61 & 64.16 \\
& SPINE~\cite{helmy2026spine}       & \underline{63.20} & 71.10 & 70.70 & 77.40 & 94.80 & 66.90 & 64.00 & 64.90 & 49.30 & 49.20 & 46.70 \\
\cmidrule(lr){2-13}
& \textbf{GateSPINE}                & \textbf{65.67} & \textbf{79.39} & \textbf{79.27} & \textbf{87.50} & \textbf{96.09} & 83.91 & \textbf{87.95} & \textbf{85.88} & 70.49 & \textbf{74.36} & \textbf{72.21} \\
\midrule
\multirow{5}{*}{\rotatebox{90}{Lumbar}}
& CT2Rep~\cite{hamamci2024ct2rep}   & \underline{20.42} & \underline{49.67} & \underline{47.76} & \underline{47.40} & \underline{90.20} & 55.66 & \underline{31.22} & 40.00 & 13.61 & 11.14 & 12.19 \\
& Dia-LLaMA~\cite{chen2025diallama} & 8.49 & 33.74 & 30.81 & 29.95 & 87.32 & 56.62 & 8.72 & 15.12 & 15.40 & 3.29 & 5.40 \\
& Med3DVLM~\cite{xin2026med3dvlm}   & 9.04 & 34.38 & 29.38 & 28.52 & 86.99 & 58.55 & 30.96 & 40.53 & 30.46 & 17.54 & 21.87 \\
& SPINE~\cite{helmy2026spine}       & 12.72 & 40.45 & 36.63 & 35.64 & 88.94 & \underline{68.54} & 30.81 & \underline{42.51} & \textbf{45.26} & \underline{19.24} & \underline{24.18} \\
\cmidrule(lr){2-13}
& \textbf{GateSPINE}                & \textbf{23.73} & \textbf{51.95} & \textbf{48.65} & \textbf{48.07} & \textbf{90.62} & \textbf{69.59} & \textbf{69.23} & \textbf{69.41} & \underline{37.69} & \textbf{34.65} & \textbf{33.39} \\
\hline\hline
\end{tabular}
\end{table}
On PhenikaaMec, GateSPINE achieves the best clinical entity scores and leads on ROUGE and BERTScore (Table~\ref{tab:result_main}). Its clinical gain comes mainly from higher recall (Micro-R 67.35 vs.\ 52.05 for Dia-LLaMA) at comparable precision, so it misses fewer true findings. Dia-LLaMA scores higher on BLEU-4 and METEOR but is much weaker on clinical entities. Across the nine findings (Fig.~\ref{fig:pathology_radar}), GateSPINE leads on seven of nine, with the clearest margins over SPINE on disc degeneration and osteophytes; it trails only on disc bulge, a naming bias rather than a miss: of 29 reference bulges it calls 24 a herniation and omits only 4. Since bulge and herniation are separate labels on a clinical continuum, this lowers Macro-F1 more than Micro-F1. On SPIDER, GateSPINE leads on all NLG metrics and on both F1 scores, with the largest margin on Macro-F1 ($+7.2$). Med3DVLM attains higher precision but far lower recall, describing fewer findings overall. Because the references here are GPT-4-generated from grading labels rather than written by radiologists, they are highly templated; absolute scores on SPIDER are inflated relative to the other two datasets and should be read only within this benchmark. On the public Lumbar dataset, GateSPINE is again best on both metric groups, above CT2Rep and SPINE. Macro-F1 stays low because every model scores zero on osteophytes and disc height loss, and the short segmentation-style references make both metric groups harder to satisfy.
 \subsection{Ablation Studies and Analyses} \label{sect:ablation}


 \noindent\textbf{Contribution of MRI Sequences.}
We assess each sequence by turning sequences on and off on PhenikaaMec and Lumbar, as shown in Table~\ref{tab:ablation_seq}. A single sequence is not enough on either dataset. On PhenikaaMec, sagittal T2 is the best single view, since most degenerative findings are clearer on T2-weighted images, and sagittal T1 is the weakest. On Lumbar the order flips: axial T2 is strongest, matching a stenosis-focused cohort where the axial plane best shows canal narrowing. Adding the axial view on top of the fused sagittal T1 and T2 gives a large jump in clinical entity scores on both datasets. The full three-sequence setup is best on every metric, and the gated integration of the axial plane drives most of the clinical entity gain.

\begin{table}[!t]
\centering
\caption{Ablation on input MRI sequences. Best in \textbf{bold}, second best \underline{underlined}.}
\label{tab:ablation_seq}
\scriptsize
\setlength{\tabcolsep}{3pt}
\begin{tabular}{@{}l | c c c | c c c c c | c c@{}}
\hline\hline
& \multicolumn{3}{c}{\textbf{Sequence}} & \multicolumn{5}{c}{\textbf{NLP Metrics}} & \multicolumn{2}{c}{\textbf{CE (F1)}} \\
\cmidrule(lr){2-4} \cmidrule(lr){5-9} \cmidrule(lr){10-11}
\textbf{Dataset} & \textbf{T1} & \textbf{T2} & \textbf{Ax} & \textbf{BL-4} & \textbf{RG-1} & \textbf{RG-L} & \textbf{MTR} & \textbf{BERT} & \textbf{Micro} & \textbf{Macro} \\
\midrule
\multirow{5}{*}{PhenikaaMec}
& \checkmark & & & 19.69 & 54.12 & 39.20 & 40.81 & 90.30 & 31.12 & 17.53 \\
& & \checkmark & & 21.81 & 54.36 & 41.16 & 42.10 & 90.18 & \underline{56.25} & \underline{46.50} \\
& & & \checkmark & 22.11 & \underline{55.64} & 42.23 & \underline{44.53} & 90.27 & 48.37 & 39.77 \\
& \checkmark & \checkmark & & \underline{23.50} & 55.52 & \underline{42.63} & 43.90 & \underline{90.43} & 49.51 & 45.16 \\
& \checkmark & \checkmark & \checkmark & \textbf{26.46} & \textbf{60.24} & \textbf{46.15} & \textbf{48.63} & \textbf{91.16} & \textbf{71.12} & \textbf{60.50} \\
\midrule
\multirow{5}{*}{Lumbar}
& \checkmark & & & 10.27 & 32.97 & 30.52 & 32.04 & 88.41 & 5.03 & 3.49 \\
& & \checkmark & & 20.91 & 43.03 & 38.80 & 42.60 & \underline{90.42} & 52.70 & 17.01 \\
& & & \checkmark & \underline{21.09} & \underline{48.22} & \underline{44.15} & \underline{45.06} & 89.70 & \underline{62.76} & 25.79 \\
& \checkmark & \checkmark & & 20.47 & 45.47 & 39.95 & 43.18 & 89.15 & 54.04 & \underline{26.44} \\
& \checkmark & \checkmark & \checkmark & \textbf{23.73} & \textbf{51.95} & \textbf{48.65} & \textbf{48.07} & \textbf{90.62} & \textbf{69.41} & \textbf{33.39} \\
\hline\hline
\end{tabular}
\end{table}

 \noindent\textbf{Cross-View Fusion Mechanism.}
Table~\ref{tab:ablation_fusion} compares three cross-view fusion methods. We first expected the uncertainty-aware dynamic fusion to perform best, but the simple gated fusion outperforms both concatenation and dynamic fusion. This holds on both PhenikaaMec and Lumbar. We attribute this to the gate acting as a stable, content-dependent selector that is easier to train on small medical data than a larger dynamic weighting module. 

\begin{table}[!t]
\centering
\caption{Ablation on the cross-view fusion mechanism (rows 1--3, MedGemma decoder) and on the language decoder (rows 3--4, gate fusion; MedGemma 4B vs.\ Phi-3-mini 3.8B). Best in \textbf{bold}, second best \underline{underlined} among fusion variants.}
\label{tab:ablation_fusion}
\scriptsize
\setlength{\tabcolsep}{2.6pt}
\begin{tabular}{@{}l l l | c c c c c | c c@{}}
\hline\hline
\multirow{2}{*}{\textbf{Dataset}} & \multirow{2}{*}{\textbf{Fusion}} & \multirow{2}{*}{\textbf{Decoder}} & \multicolumn{5}{c}{\textbf{NLP Metrics}} & \multicolumn{2}{c}{\textbf{CE (F1)}} \\
\cmidrule(lr){4-8} \cmidrule(lr){9-10}
& & & \textbf{BL-4} & \textbf{RG-1} & \textbf{RG-L} & \textbf{MTR} & \textbf{BERT} & \textbf{Micro} & \textbf{Macro} \\
\midrule
\multirow{4}{*}{PhenikaaMec}
& Concatenation & MedGemma & 23.61 & \underline{56.80} & 43.58 & 44.67 & 90.59 & 53.11 & 46.65 \\
& Dynamic       & MedGemma & \underline{24.37} & 56.60 & \underline{44.31} & \underline{45.37} & \underline{90.73} & \underline{63.01} & \underline{53.46} \\
& \textbf{Gate} & MedGemma & \textbf{26.46} & \textbf{60.24} & \textbf{46.15} & \textbf{48.63} & \textbf{91.16} & \textbf{71.12} & \textbf{60.50} \\
\cmidrule(lr){2-10}
& Gate          & Phi-3-mini & 20.94 & 53.47 & 39.28 & 41.51 & 89.93 & 56.32 & 46.89 \\
\midrule
\multirow{4}{*}{Lumbar}
& Concatenation & MedGemma & 18.56 & 44.25 & 40.79 & 40.56 & 89.59 & 53.90 & 23.47 \\
& Dynamic       & MedGemma & \underline{21.17} & \underline{47.35} & \underline{43.01} & \underline{44.67} & \underline{89.69} & \underline{62.32} & \underline{32.42} \\
& \textbf{Gate} & MedGemma & \textbf{23.73} & \textbf{51.95} & \textbf{48.65} & \textbf{48.07} & \textbf{90.62} & \textbf{69.41} & \textbf{33.39} \\
\cmidrule(lr){2-10}
& Gate          & Phi-3-mini & 18.35 & 47.10 & 43.45 & 43.08 & 89.70 & 59.83 & 24.68 \\
\hline\hline
\end{tabular}
\end{table}

 \noindent\textbf{LLM as the Language Decoder.} To isolate the decoder's role, we keep the 3D visual architecture, gated fusion, and projector fixed and swap only the LLM decoder, as shown in Table~\ref{tab:ablation_fusion}. MedGemma outperforms Phi-3 on every metric. The gap is small on language metrics but much larger on clinical efficacy. We attribute this to MedGemma's biomedical pre-training, which turns the same features into more accurate findings.

\begin{figure*}[!t]
\centering
\includegraphics[width=0.9\textwidth,trim={36 203.76 38.16 41.04},clip]{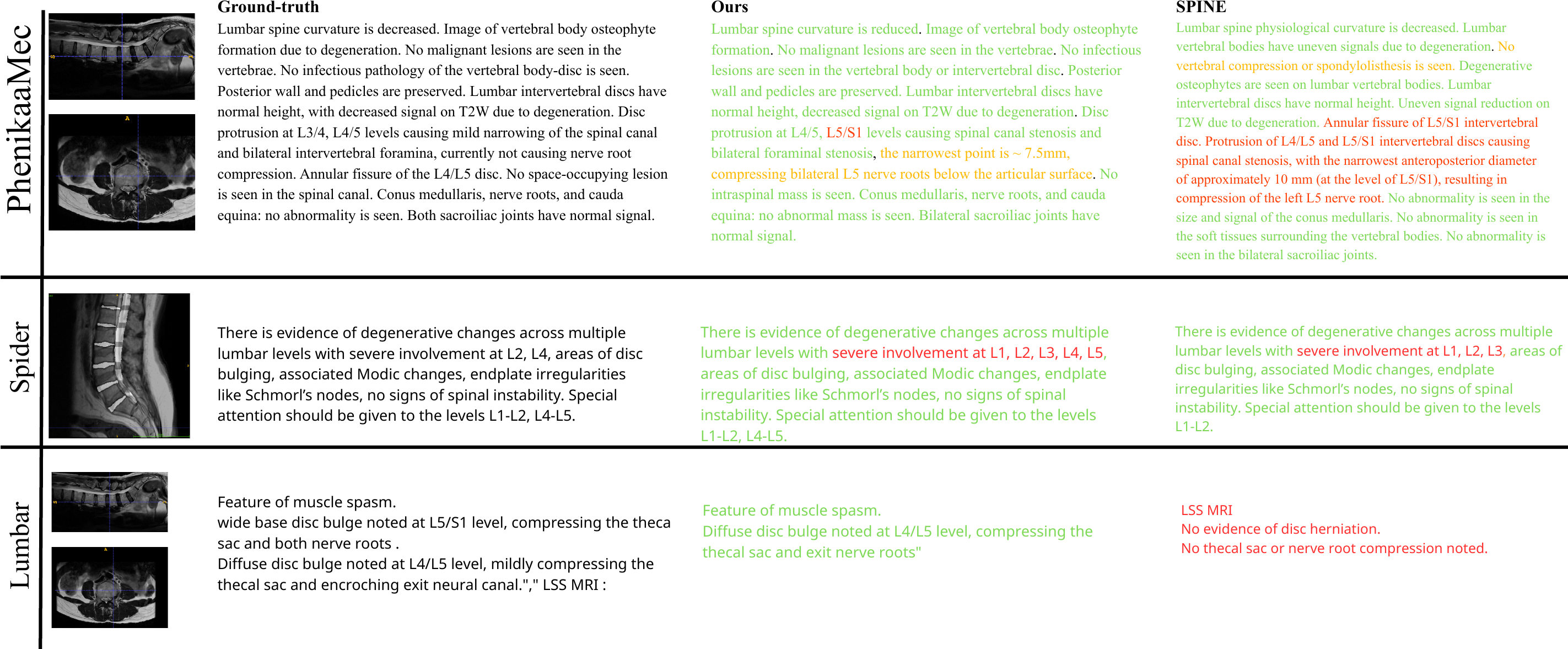}
\caption{Generated report comparison: Ground-truth, Ours, and SPINE, one case per dataset.}
\label{fig:compare_report}
\end{figure*}

 \noindent\textbf{Analysis of Generated Reports.}
Figure~\ref{fig:compare_report} shows one representative case from each dataset, comparing generated and ground-truth reports and judged on abnormality type, location, and severity. Across all three datasets, GateSPINE recovers most major abnormalities with fewer hallucinated and missed findings than SPINE, which both misses findings and adds irrelevant text. The remaining GateSPINE errors are concentrated in the vertebral level and in lesion severity, which we target in future work.

\section{Conclusions} \label{sect:conclusion}

In this paper, we propose \textbf{GateSPINE}, a unified vision--language framework for lumbar spine MRI report generation. Unlike existing methods that do not fully utilize the complementary information from different MRI views, GateSPINE introduces a lightweight gated cross-view fusion module to adaptively combine sagittal and axial representations. The proposed module learns how much information to preserve from each view for every feature channel, enabling the model to better exploit complementary information across MRI sequences and imaging planes. On one private and two public lumbar MRI datasets, GateSPINE achieves the best performance on clinical-efficacy metrics while remaining competitive on language-generation metrics. Furthermore, ablation studies show that the gated cross-view fusion module outperforms existing fusion strategies. 

One limitation of the current framework is that it mainly relies on global image representations, which can cause the model to occasionally miss subtle lesions or describe their anatomical locations inaccurately, as illustrated in Fig.~\ref{fig:compare_report}. In future work, we plan to incorporate finer-grained anatomical information, such as disc-level segmentation or lesion detection, to improve localization accuracy and further enhance report quality.

\section*{Acknowledgements}
This research was partially funded by Vingroup Innovation Foundation (VINIF) under project code VINIF.2026.DA036. This research was also supported by the Ben Dam Me Award Fund, the Vietnam Young Talent Support Fund, and the Number One Brand, Tan Hiep Phat Group.

\bibliographystyle{splncs04}
\bibliography{references}
\end{document}